\documentclass{ieeeaccess}
\usepackage{cite}
\usepackage{amsmath,amssymb,amsfonts}
\usepackage{algorithmic}
\usepackage{graphicx}
\usepackage{textcomp}
\def\BibTeX{{\rm B\kern-.05em{\sc i\kern-.025em b}\kern-.08em
    T\kern-.1667em\lower.7ex\hbox{E}\kern-.125emX}}

\usepackage{amssymb}
\usepackage{amsmath}
\usepackage{hhline}
\usepackage{adjustbox}
\usepackage{multirow}
\usepackage{tabularx}
\usepackage{comment}
\usepackage{caption}
\usepackage{xspace}
\usepackage{xcolor}
\makeatletter
\DeclareRobustCommand\onedot{\futurelet\@let@token\@onedot}
\def\@onedot{\ifx\@let@token.\else.\null\fi\xspace}

\def\eg{\emph{e.g}\onedot} 
\def\ie{\emph{i.e}\onedot}

\makeatother

\begin{document}
\history{Date of publication xxxx 00, 0000, date of current version xxxx 00, 0000.}
\doi{10.1109/ACCESS.2017.DOI}

\title{Projection-based Point Convolution for Efficient Point Cloud Segmentation}
\author{\uppercase{Pyunghwan Ahn}\authorrefmark{1}, \uppercase{Juyoung Yang}\authorrefmark{1}, \uppercase{Eojindl Yi}\authorrefmark{1}, \uppercase{Chanho Lee}\authorrefmark{1}, and \uppercase{Junmo Kim}\authorrefmark{1},
\IEEEmembership{Member, IEEE}}
\address[1]{School of Electrical Engineering, Korea Advanced Institute of Science and Technology, Daejeon 34141, South Korea}

\tfootnote{This research was supported by the Engineering Research Center Program through the National Research Foundation of Korea (NRF) funded by the Korean Government MSIT (NRF-2018R1A5A1059921).}

\markboth
{Author \headeretal: Preparation of Papers for IEEE TRANSACTIONS and JOURNALS}
{Author \headeretal: Preparation of Papers for IEEE TRANSACTIONS and JOURNALS}

\corresp{Corresponding author: Junmo Kim (e-mail: junmo.kim@kaist.ac.kr).}

\begin{abstract}
Understanding point cloud has recently gained huge interests following the development of 3D scanning devices and the accumulation of large-scale 3D data. Most point cloud processing algorithms can be classified as either point-based or voxel-based methods, both of which have severe limitations in processing time or memory, or both. To overcome these limitations, we propose Projection-based Point Convolution (PPConv), a point convolutional module that uses 2D convolutions and multi-layer perceptrons (MLPs) as its components. In PPConv, point features are processed through two branches: point branch and projection branch. Point branch consists of MLPs, while projection branch transforms point features into a 2D feature map and then apply 2D convolutions. As PPConv does not use point-based or voxel-based convolutions, it has advantages in fast point cloud processing. When combined with a learnable projection and effective feature fusion strategy, PPConv achieves superior efficiency compared to state-of-the-art methods, even with a simple architecture based on PointNet++. We demonstrate the efficiency of PPConv in terms of the trade-off between inference time and segmentation performance. The experimental results on S3DIS and ShapeNetPart show that PPConv is the most efficient method among the compared ones. The code is available at github.com/pahn04/PPConv.
\end{abstract}


\titlepgskip=-15pt

\maketitle

\section{Introduction}
\label{sec:introduction}

Recent developments in 3D scanning devices have incorporated a great amount of 3D data into machine vision applications, such as robotics, autonomous vehicles, and VR/AR. Point cloud is a popular type of data with 3D geometry, and thus it is significantly beneficial to have a reliable autonomous point cloud processing system for better 3D perception. Although computer vision algorithms have achieved remarkable improvements in image recognition, 3-dimensional data is naturally different from images that have only 2 spatial dimensions. Thus, researchers have studied novel methods to deal with 3D data effectively.

As convolutional neural networks (CNNs) have been proved to be effective in image recognition, researchers have attempted to extend the use of CNNs to 3D data processing. Several voxel-based methods~\cite{maturana2015voxnet, wang2017cnn, graham20183d} applied CNNs to 3D data, but this type of approach usually suffered from computational complexity of 3D convolutions. Higher voxel resolution led to a rapid increase in memory requirements and computation time, whereas lower resolution caused larger quantization errors. Alternatively, point-based methods~\cite{qi2017pointnet,qi2017pointnet++,li2018pointcnn,xu2018spidercnn,wang2019dynamic} could alleviate the exhaustive memory requirements, by allocating memory that is linearly proportional to the number of points. However, during local information aggregation, frequent neighbor search and dynamic kernel computation led to delayed latency due to the irregular memory access and additional alignment computation, as pointed out in \cite{liu2019point}.

In this study, we aim to design a point convolutional module that does not have the limitations of 3D convolutions and point-based convolutions. Therefore, we choose 2D convolutional blocks as the fundamental operator. Even though projection-based methods for 3D data have been suggested before~\cite{su2015multi,roveri2018network,li2020end,yang2020pbp}, they showed degraded performance due to the nature of 2D convolutions, which distorts the 3D geometry and thus is not as expressive as 3D operators on 3D data. We intend to tackle a few problems that projection-based methods usually face. First, an N$\times$C point cloud needs to be transformed into a H$\times$W$\times$C feature map to be processed with 2D convolutions. In naive projection methods, \eg, removing one of the coordinates, information loss is inevitable due to the dimension reduction. Next, when projected features are processed through separate CNNs and merged at the final stage, intermediate features are not aware of context information from other projections, which limits the advantage of multi-view projection.

To this end, we propose Projection-based Point Convolution (PPConv), a point convolutional module for point cloud processing with 2D convolutions. In PPConv, two separate branches process point features: point branch applies MLPs to point features and projection branch uses 2D convolutions. In the projection branch, point clouds are projected through PointNet-based learnable projection, which minimizes the information loss during the dimension reduction. Furthermore, features from each branch are fused inside the convolutional module, so that the output of PPConv is always the aggregated information of all the branch features. For the feature fusion, we propose Importance-Weighted Fusion and Context-Aware Fusion modules that enables effective fusion of multiple features. By integrating all these modules, PPConv maximizes the effectiveness of 2D convolutions on point cloud processing.

We demonstrate performance and computational efficiency of PPConv by means of comparison with several state-of-the-art methods. To strictly measure the effect of convolutional modules, we use a backbone of PointNet++~\cite{qi2017pointnet++} with architecture hyperparameters used in \cite{liu2019point}, and only replace the convolutional modules by PPConv. The experimental results on S3DIS~\cite{armeni20163d} indoor scene segmentation demonstrate that PPConv shows superior efficiency among the compared methods, in terms of the trade-off between inference time and segmentation performance. Furthermore, PPConv also achieves comparable performance to the state-of-the-art on ShapeNetPart~\cite{yi2016scalable}.

Contributions of this paper are as follows:
\begin{itemize}
\item We propose the 2D convolution-based point convolutional module, which can be used as a building block of a point-based network for point cloud processing.
\item We propose novel feature fusion modules which can effectively fuse point features generated by different types of convolutional operations.
\item We compare the inference time of state-of-the-art methods and some representative studies on the same hardware environment, using their publicly available implementations, and show that PPConv is one of the most efficient modules in terms of the trade-off between inference time and performance.
\end{itemize}

\section{Related works}\label{sec:related_works}

\textbf{Voxel-based methods} use voxel representations for 3D data processing. VoxNet~\cite{maturana2015voxnet} used 3D CNNs on 3D data, such as raw LiDAR point clouds, RGB-D data, and CAD models, after transforming them into voxel representations. To address the computational burden of 3D convolutions, octree-based method~\cite{wang2017cnn} and sparse 3D convolution~\cite{graham20183d} were proposed, both of which reduced redundant computations at unoccupied spaces. Another study~\cite{choy20194d} proposed optimized sparse convolution for processing point cloud with varying density.

\textbf{Point-based methods} have recently focused on hierarchical feature generation on point clouds, following the pioneering work of PointNet~\cite{qi2017pointnet}. PointNet++~\cite{qi2017pointnet++} used PointNet on local point sets to extract local features, while some other works~\cite{li2018pointcnn, xu2018spidercnn, liu2019relation, duan2019structural} attempted to infer weights from the relative location or local structure of the points. Local projection was proposed in \cite{tatarchenko2018tangent,lin2020fpconv}, in which 2D convolutions were applied to extract local features. Various point-based convolutional operations~\cite{atzmon2018point, zhao2019pointweb, thomas2019kpconv, mao2019interpolated, xu2021paconv, fan2021scf, qiu2021semantic, wu2019pointconv, liu2019densepoint} were recently proposed to properly capture local context of point clouds. Construction of graph was also suggested in other works~\cite{landrieu2018large,wang2019dynamic,wang2018local,lin2021learning,yi2017syncspeccnn} for effective point cloud processing. Some studies focused on the speed of point-based networks, by voxel-based neighbor search~\cite{xu2020grid} or random down-sampling~\cite{hu2020randla}. Other recent studies attempted to improve segmentation by predicting labels in each local region~\cite{gong2021omni} or by matching categories with neighboring points~\cite{lu2021cga}.

Several other studies have attempted to combine the advantages of voxel-based and point-based methods. In PVCNN~\cite{liu2019point}, point-wise MLP was used along with 3D convolutions on coarsely voxelized data.
MLP extracted the individual point features, while 3D convolutions aggregated local information from adjacent voxels. These two features were added to produce the output of a PVConv module. The aid of MLP alleviated the requirement of high voxel resolution, because MLPs could focus on fine-grained features while voxel features captured larger contextual features. Lower voxel resolution led to significant improvement in computational efficiency, which is the main contribution of PVCNN. Our research is highly motivated by PVCNN and mainly compared with it.
Following PVCNN, Sparse Point-Voxel Convolution~\cite{tang2020searching} was suggested to process sparse point cloud with PVConv, and FusionNet~\cite{zhang2020deep} was proposed to extract local features with an improved voxel feature aggregation module.

In \textbf{Projection-based methods}, 2D CNNs have been used in 3D shape recognition or scene analysis. CNNs have been applied to rendered images~\cite{su2015multi} or depth images~\cite{roveri2018network} transformed from 3D data. Projection-based methods have also been proposed for object detection~\cite{lang2019pointpillars, zhou2020end, wang2020pillar}, scene segmentation~\cite{lawin2017deep,yang2020pbp}, registration~\cite{li2020end}, and 3D reconstruction~\cite{peng2020convolutional, lionar2021dynamic}. Furthermore, projection onto higher-dimensional lattices was suggested for effective feature aggregation~\cite{su2018splatnet}.

\section{Motivation}
\label{sec:motivation}

\begin{figure*}[ht!]
    \centering
    \includegraphics[width=\textwidth]{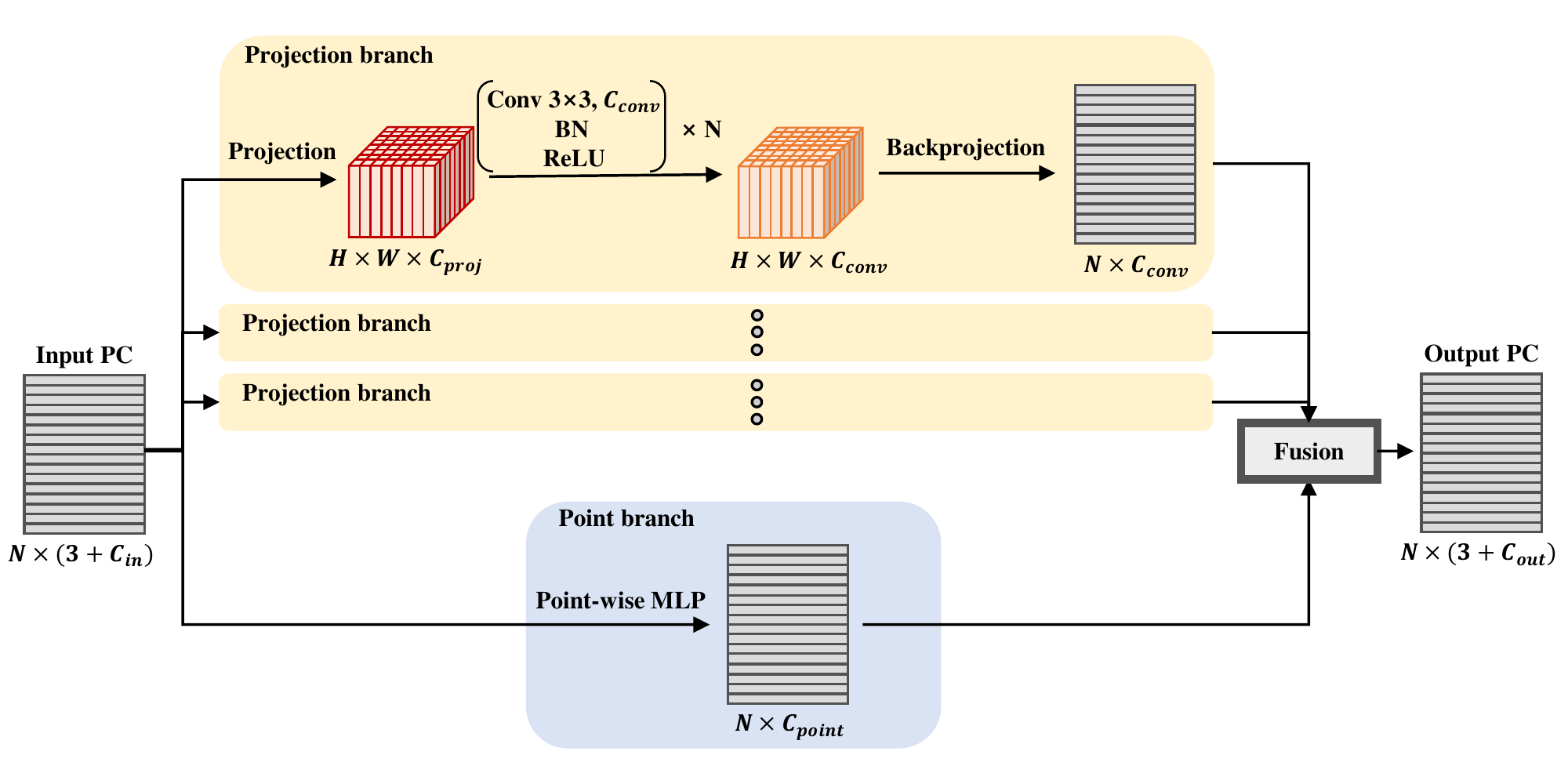}
    \caption{Description of PPConv module. The input point cloud is processed through projection branch and point branch, and then fused together to produce the final output point-wise features.}
    \label{fig:method}
\end{figure*}

Most of the recent point cloud processing networks use either voxel-based 3D convolutions or point-based convolutional operations, as summarized in section~\ref{sec:related_works}. However, each group of methods has critical limitations as pointed out in \cite{liu2019point}. In voxel-based methods, memory usage and computational cost grow cubically with respect to the voxel resolution, which is considered to be a severe problem in scalablity. To alleviate this problem, some recent studies~\cite{choy20194d, tang2020searching, graham20183d, zhang2020deep} proposed to use sparse 3D convolution, which helps reduce redundant computations and thus enables to use small voxels in large scenes. In this paper, we take a different strategy to reduce computation by not using 3D convolutions at all, but instead processing point clouds with only 1D and 2D convolutions.

Point-based methods~\cite{li2018pointcnn, xu2018spidercnn, thomas2019kpconv} suffer from delayed processing time caused by the additional computations needed for nearest neighbor search or kernel weight interpolation. In point clouds, the neighboring points are not saved in the memory contiguously: local point grouping operations require either radius search or k-nearest neighbor search. Another problem is that the input point locations vary for every sample, so the kernel points must be aligned with them before computing features. Based on these observations, we aim to use grid-based convolutions as the core operation to avoid inefficient computations. Thus, our design goal can be summarized as to process 3D point clouds with 1D and 2D operations on a grid structure.

\section{Projection-based Point Convolution}
\label{sec:method}

In this section, we introduce Projection-based Point Convolution (PPConv) module. The overall framework is depicted in Fig.~\ref{fig:method}. As a point-based convolutional method, PPConv module takes a point cloud $P = \{p_1, p_2, \cdots, p_{N}\} \in \mathbb{R}^{N\times(3+C_{in})}$ as input and outputs the transformed point cloud $P' \in \mathbb{R}^{N'\times(3+C_{out})}$. Each point $p_i$ contains its point coordinates $c_i$ and point features $f_i$, \ie, $p_i = (c_i, f_i)$. In the first layer of the network, the input may be only the 3D coordinates of each point, or may have extra features such as RGB values or normalized coordinates ($C_{in} = 6$ when RGB values and normalized 3D coordinates are used). In the other layers, each point has its coordinates $c_i$ and features $f_i$, which are the output of the previous layer ($C^l_{in} = C^{l-1}_{out}$, where $l$ is the layer index).
In PPConv, $P$ is processed through two separate pathways: the point branch and the projection branch. In the point branch, MLP transforms each point feature individually. In the projection branch, the point cloud is projected onto a 2D plane and subsequently processed using 2D convolutions. Then, the feature map is backprojected to the point space and combined with the other features through fusion module. Each branch is further explained in the following subsections.

\subsection{Point branch}
\label{sec:point_branch}

In point branch, point-wise feature transformation is performed through a single-layer MLP, followed by batch normalization and ReLU:
\begin{equation}
    f^{point}_k = ReLU(BN(MLP_{point}(f_k))),
\end{equation}
where $f^{point}_k \in \mathbb{R}^{C_{point}}$. For the experiments presented in this paper, we use $C_{point} = C_{out}/2$, where $C_{out}$ is the output channel of PPConv module. As MLP transforms individual point features, this branch is able to extract features that differ from point to point.

\subsection{Projection branch}
\label{sec:projection_branch}

As described in Fig.~\ref{fig:method}, the projection-based feature extraction can be divided into three steps: projection, 2D convolutional module, and backprojection. Projection aggregates information along the projection axis, and then 2D convolutional module aggregates information along the other two axes. In this subsection, we provide the fundamental framework, and each module can be selected among several candidates depending on the characteristics or complexity of the target data.

\subsubsection{Projection}
\label{sec:projection}

Projection of a point cloud onto a 2D plane can be performed using several different methods. The projection plane is basically divided into grid, and then the key problem is how the points inside each cell can be aggregated into a single feature vector. The simplest way is to average features of points in the same cell. To further reflect the detailed point locations, bilinear interpolation can be applied to each point feature. During bilinear interpolation, a point feature affects four nearest grid cells. This method is useful in cases where most grid cells include only a few points.

If more points belong to a grid cell, more effective methods are required to avoid information loss during projection. In some previous studies~\cite{lang2019pointpillars,zhou2018voxelnet}, a mini-PointNet structure is employed to aggregate multiple point features inside a pillar or a voxel. Following these works, we use PointNet-based projection to aggregate information of multiple points in a grid cell. Since PointNet aggregates information of arbitrary number of points into a single feature vector, it can produce a fixed-length feature for each grid cell. The PointNet-based projection consists of MLPs and max-pooling in each grid cell:
\begin{equation}\label{eq:2}
\begin{aligned}
    f^{proj}_{(i,j)} = MaxPo&ol_{K(i,j)}(MLP_{proj}(f_k, f^{aug}_k)),\\
    &\pi(k) = (i,j),
\end{aligned}
\end{equation}
where $\pi(k) = (i,j)$ is point-to-grid index mapping function, which indicates that the $(i,j)$-th grid cell is where the point $p_k$ falls into. $K(i,j)$ denotes the index set of all the points that fall into the grid cell $(i,j)$.
As in \cite{lang2019pointpillars}, the input features are augmented with the relative coordinates to the arithmetic mean of coordinates of the points inside the grid cell ($x_c, y_c, z_c$) and the relative location to the grid center ($x_p, y_p$ in case of z-axis projection). After projection, a 2D feature map is constructed by stacking a feature vector for every grid cell. The feature vector of an empty cell with no points is filled with zero.

\subsubsection{2D Convolutional module}
\label{sec:2Dconv}

After projection, the 2D convolutional modules are used to transform the projected feature map. Since a 2D feature map has been produced through projection, any techniques used in conjunction with 2D convolution can be seamlessly integrated into PPConv. In this paper, we use a basic residual block~\cite{he2016deep} with 2 convolution layers, each followed by batch normalization~\cite{ioffe2015batch} and leaky ReLU. The Squeeze-and-Excitation (SE) module~\cite{hu2018squeeze} is incorporated to further improve training.
\begin{equation}
    f^{conv}_{(i,j)} = SEResBlock(f^{proj}_{(i,j)}).
\end{equation}

\subsubsection{Backprojection}
\label{sec:backprojection}

\begin{figure*}[ht!]
    \centering
    \includegraphics[width=\textwidth]{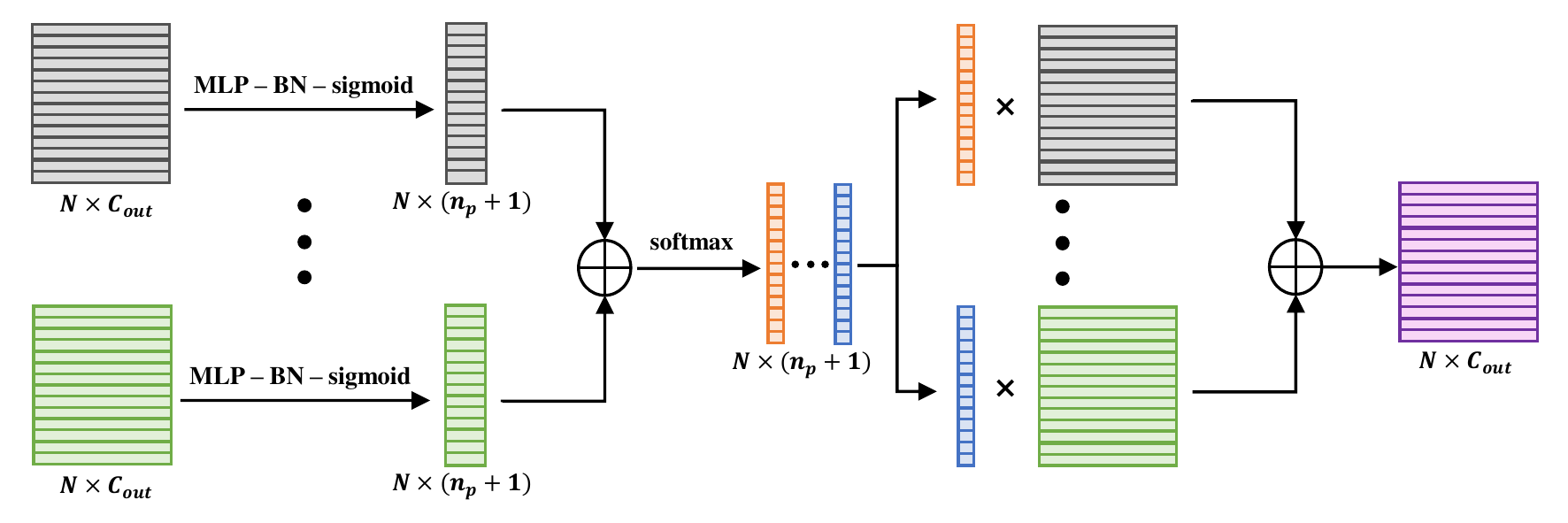}
    \caption{Importance-Weighted Fusion (IWF) module.}
    \label{fig:iwf}
\end{figure*}

\begin{figure*}[ht!]
    \centering
    \includegraphics[width=\textwidth]{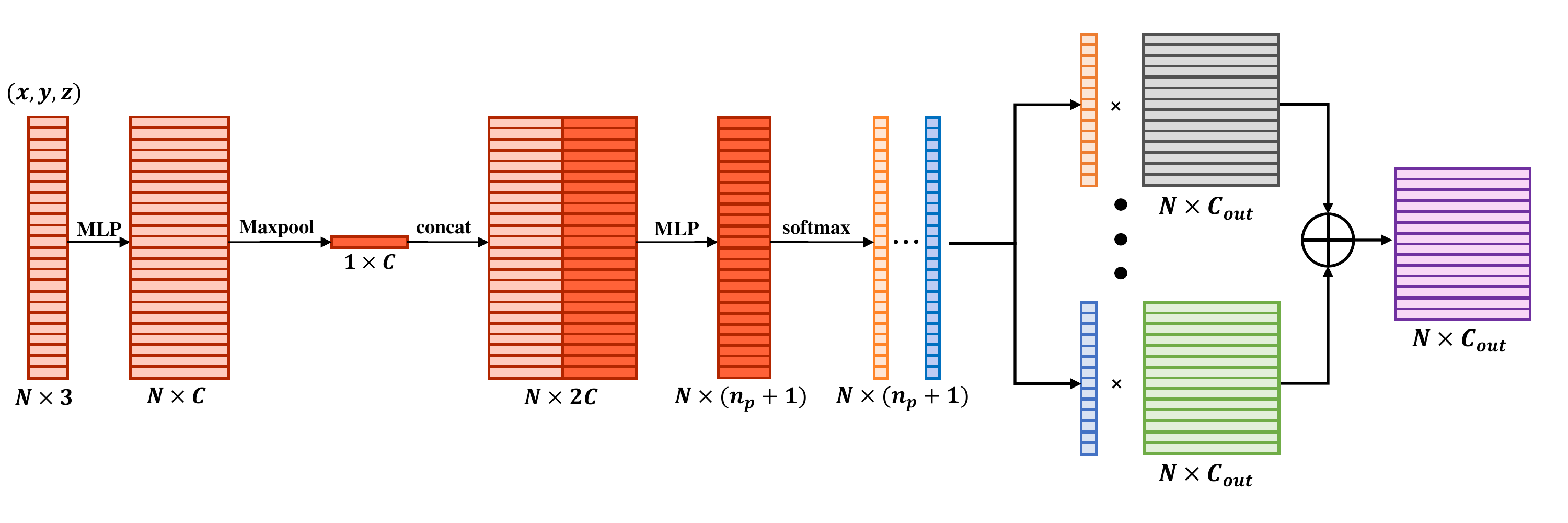}
    \caption{Context-Aware Fusion (CAF) module.}
    \label{fig:caf}
\end{figure*}

After being transformed by 2D convolutions, feature maps are backprojected to the original point locations.
Backprojection is performed by calculating each point feature based on the features of the 2D grid cell that the point was projected to. As shown in Fig.~\ref{fig:method}, backprojection transforms 2D feature map ($H \times W \times C_{conv}$) into point features ($N \times C_{conv}$); thus a mapping function between grid cell indices $H \times W$ and point indices $N$ is needed. A natural choice would be reusing the mapping function used in projection: $\pi(\cdot)$ in (\ref{eq:2}). Based on this mapping function, backprojection can be performed through several possible methods, the simplest of which is the nearest neighbor assignment. In nearest neighbor assignment, each grid cell feature (with channel size of $C_{conv}$) of the 2D feature map is assigned to every point that corresponds to that cell.
However, this forces the points that share a grid cell to have exactly the same backprojected features. To efficiently reflect the contribution of each grid cell feature to the individual point, the distance between a point and the center of the corresponding grid cell is used to calculate the weights:
\begin{equation}\label{eq:4}
\begin{aligned}
    &f^{backproj}_k = \;w_k \times f^{conv}_{(i,j)},\\
    w_k &= \;1-|c_k[a_1,a_2]-m(i,j)|_1,&
\end{aligned}
\end{equation}
where $\pi(k) = (i,j)$ is the point-to-grid index mapping function as in (\ref{eq:2}), $a_i$ are the axis indices except the projection axis, $m(i,j)$ is the center of grid cell $(i,j)$. $c_k$ are the point coordinates of point $p_k$; thus $c_k[a_1,a_2]$ is the location of $p_k$ on the projection plane (\eg ($x_k$, $y_k$) in case of $z$-axis projection).

During experiments, we further investigated various backprojection modules. However, the effect of fusion module was much more significant and thus various backprojection methods did not show noticeable differences. Thus, we use a simple module presented in (\ref{eq:4}) and focus on the fusion module.

\subsection{Feature fusion}
\label{sec:feature_fusion}

Once each branch has processed features through different operations, they need to be effectively merged to leverage the advantages of different branches. In PPConv, the basic feature aggregation is performed by concatenating two features, one from point branch and the other from projection branches, and then applying a single-layer MLP. In case of PPConv with multiple projection branches, the backprojected features are added before being combined with point branch features. Throughout the experiments, we use $C_{proj} = C_{conv} = C_{out}/2$ so that the concatenated feature has the channel dimension of $C_{out}$.

Furthermore, we propose two different methods of learning-based weighted aggregation of multiple features. Here, we investigate the fusion of $n_p+1$ features, where $n_p$ is the number of projection branches and $+1$ is for the point branch. The first method is Importance-Weighted Fusion (IWF), which is depicted in Fig.~\ref{fig:iwf}. Features extracted from each branch are transformed through a single-layer MLP and sigmoid function to produce an $N\times(n_p+1)$ matrix. Then, these matrices are summed to form one $N\times(n_p+1)$ matrix and then softmax is applied to each row of this matrix. Each row, which corresponds to one of the $N$ points, represents the fusion weights for each of the $n_p+1$ features of that point. Finally, the corresponding weight is multiplied to each feature and summed to compute the output of PPConv module. This approach enables the network to learn how to fuse features by computing weights depending on the importance of each feature.

Another proposal of feature fusion strategy is Context-Aware Fusion (CAF), described in Fig.~\ref{fig:caf}. In this module, the point coordinates ($N \times 3$) are fed into a mini-PointNet structure to produce $N \times (n_p+1)$ matrix. Then, the weights, after applying softmax, are used in the same way as in IWF module. CAF module uses the geometry of the input point cloud to produce the weight matrix. Intuitively, CAF can be considered to be deciding which features are more informative for each point based on the shape of current input point cloud. Furthermore, CAF includes a max-pooling operation over the point dimension, which allows the attention weights to reflect the global context.

\subsection{Discussion}
\label{sec:discussion}

In this subsection, we explain the difference of PPConv from other methods in detail. Primarily, following our design goal explained in Section~\ref{sec:motivation}, PPConv utilizes 2D convolutions to efficiently extract local features, unlike voxel-based or point-based methods. Projection-based point convolutional networks proposed in some studies~\cite{tatarchenko2018tangent,lin2020fpconv} require neighbor search and local point grouping, while in PPConv, they are efficiently carried out using grid cell index. Comparing with projection-based 2D CNN models~\cite{lang2019pointpillars,wang2020pillar,zhou2020end}, PPConv fuses multi-view projection features and point-wise features in every convolutional module, while existing studies focused on aggregating the final features of CNNs.

In PPConv, two feature fusion modules are proposed to aggregate features from different branches. IWF resembles attentive fusion strategies presented in recent point cloud segmentation studies~\cite{qiu2021semantic, cheng20212}. The previous works proposed to summarize each $N \times C$ feature into an $N \times 1$ vector using MLPs and then concatenate these vectors before normalizing them with softmax to calculate attention weights. In IWF, an $N \times (n_p+1)$ matrix is calculated from each $N \times C$ feature, thus enabling every feature to directly affect the weights for all the other features. Additionally, CAF module takes the point cloud geometry into account while calculating the weights, which is a novel approach to attention-based feature fusion, since other methods generally use the same features for computation of attention weights and for fusion based on those weights.

Recently, feature aggregation based on self-attention has been popularized in computer vision tasks, on both images~\cite{dosovitskiy2020image,touvron2021training} and point clouds~\cite{zhao2021point}. These attention mechanisms, called transformers, are effective in various applications but known to consume significant computational resources as they require to calculate pairwise relations of the features. Compared to transformers, the proposed feature fusion modules are much simpler and thus bring small increase of runtime. Furthermore, the intermediate features in the fusion modules have a smaller channel dimension than the features of each branch, which also brings a small increase in memory requirements. Therefore, through the proposed feature fusion modules, we can achieve efficient 3D feature extraction using 2D convolutions.

\section{Network architecture and loss}
\label{sec:network_architecture}

In order to demonstrate the effect of PPConv module, we follow the PointNet++~\cite{qi2017pointnet++} architecture as used in PVCNN~\cite{liu2019point}. In PointNet++, Set Abstraction (SA) module aggregates the local geometry of each sampled point through a local PointNet. Feature Propagation (FP) modules use MLPs to aggregate multi-level features, which consist of interpolated higher-level features and previously extracted lower-level features from the corresponding SA module. We refer the readers to PointNet++ paper~\cite{qi2017pointnet++} for further details.

We follow the layer configurations of PVCNN++ architectures to purely compare the effectiveness of the convolutional module. For PPConv, we use the grid resolution of 64 for the first layer, which is determined through ablation study (in Table~\ref{tab:ablation_grid_resolution}). Hereafter, PointNet++ architecture equipped with PPConv is denoted as PPCNN++. In Section~\ref{sec:experiment}, PPCNN++ architectures with different configurations are indicated by PPCNN++(config). For example, PPCNN++ model with 3 projection branches (x-, y-, and z-axis projection) and Importance-Weighted Fusion (IWF) module is denoted as PPCNN++(x,y,z,IWF).

Following recent state-of-the-art methods~\cite{thomas2019kpconv,choy20194d,qiu2021semantic}, we use cross entropy loss for each point. As semantic segmentation task targets for a class prediction for every input point, we simply apply the point-wise classification loss to each point and average those values to get the loss value for the input point cloud.

\section{Experiment}
\label{sec:experiment}

\subsection{Settings}
\label{sec:settings}

In this section, we present the experiments to demonstrate the efficiency of PPConv module. We performed experiments on two datasets of different scales: indoor scene segmentation on S3DIS~\cite{armeni20163d} and object part segmentation on ShapeNetPart~\cite{yi2016scalable}. Both datasets are aligned; each axis (x, y, or z) of 3D coordinates shows a consistent direction throughout the dataset. Thus, we only use the x-, y-, and z-axis projections in this paper.

We report experimental results along with the inference time of each model. For the segmentation performance, we report the best mIoU of each model for fair comparison with other works. Performances of other methods were copied from the corresponding published paper, and the inference time was measured in a fixed hardware environment using the code provided by the authors. Because runtime of each method reported in the paper may vary depending on the hardware environment, we only report those measured in our setting: inference on an NVIDIA RTX 2080 Ti GPU following warm-up and synchronization of the GPU.

\subsection{Indoor Scene Segmentation}
\label{sec:scene_seg}

\subsubsection{Dataset}
\label{sec:scene_seg_dataset}

For indoor scene segmentation, we used Stanford Large-Scale 3D Indoor Spaces (S3DIS)~\cite{armeni20163d} dataset. In S3DIS, point clouds are scattered over 271 rooms in 6 areas, and area 5 is often used for testing as it contains no region that overlaps with the other areas. Following previous works, we trained the models on areas 1 $\sim$ 4 and 6, and then tested on area 5. Each point in the dataset is assigned one out of 13 categories, including large objects (\eg, ceiling and floor) and small objects (\eg, beam and chair). We used 6 extra feature channels as the input following the majority of recent studies, which are RGB values of each point and normalized 3D coordinates with respect to the room size.

The data pre-processing and evaluation method for S3DIS experiments were borrowed from PVCNN~\cite{liu2019point} and FPConv~\cite{lin2020fpconv}. PVCNN used 8,192 points for each 1.5m $\times$ 1.5m block as input, while FPConv used 14,564 points in a 2.0m $\times$ 2.0m block. Furthermore, the blocks are divided in the pre-processing step in PVCNN and the same block is used for the whole training procedure, while in FPConv, the block location is randomly sampled on the fly. We evaluate PPCNN++ with these two data pre-processing methods. In this paper, we denote PPCNN++ models trained under PVCNN's experimental protocol as ``PPCNN++ (PV)'', and those trained under FPConv's protocol as ``PPCNN++ (FP)''.

\begin{table}[h]
\centering
\caption{Network layer configurations of PPCNN++ for S3DIS experiments}
\begin{adjustbox}{max width=\columnwidth}
\begin{tabular}{|c|c|c|}
\hline
\multirow{4}{*}{SA blocks} & [32, 2, \textbf{64}] & 1024, 0.1, 32, (32, 64) \\
& [64, 3, \textbf{32}] & 256, 0.2, 32, (64, 128) \\
& [128, 3, \textbf{16}] & 64, 0.4, 32, (128, 256) \\
& None & 16, 0.8, 32, (256, 256, 512) \\ \hline
\multirow{4}{*}{FP blocks} & (256, 256) & [256, 1, 8] \\
& (256, 256) & [256, 1, \textbf{16}] \\
& (256, 128) & [128, 2, \textbf{32}] \\
& (128, 128, 64) & [64, 1, \textbf{64}] \\ \hline
\end{tabular}
\end{adjustbox}
\label{tab:arch_net_s3dis}
\end{table}

\subsubsection{Architecture and Hyperparameters}
\label{sec:scene_seg_arch_param}

As explained in Section~\ref{sec:network_architecture}, we used the same network architecture with PVCNN++, except for the grid resolution, which we define as 64 $\times$ 64 for the first layer. The layer configuration is shown in Table~\ref{tab:arch_net_s3dis}. In the table, the numbers in each set of [ ] denote output channel, number of layers, and grid resolution, while the numbers in each set of ( ) denote channel dimensions of MLP layers. The other numbers in each row of SA blocks denote number of sampled points, radius of neighborhood, and number of neighboring points to sample. In PPConv modules, we used PointNet-based projection and distance-weighted sum backprojection, which are explained in Section~\ref{sec:projection_branch}.

During testing, we adjusted the number of input points while measuring inference time for fair comparison. For the majority of methods, the number of input points can be controlled using the batch size, while other methods, such as MinkowskiNet~\cite{choy20194d} and KPConv~\cite{thomas2019kpconv}, take input with varying number of points. Thus, we set the batch size as the minimum value such that the total number of input points exceed the average number of input points of MinkowskiNet (51k). This results in the inference batch size of 7 for the model that takes 8,192 points per sample, and 4 for the model with 14,564 points per sample. For the methods with different number of input points, the batch size was adjusted so that the total number of input points do not exceed 7$\times$8192. Therefore, we ensure that the inference time of PPCNN++ was measured using the number of input points that is comparable or larger than all the other methods.

\begin{figure}[h]
    \centering
    \includegraphics[width=\columnwidth]{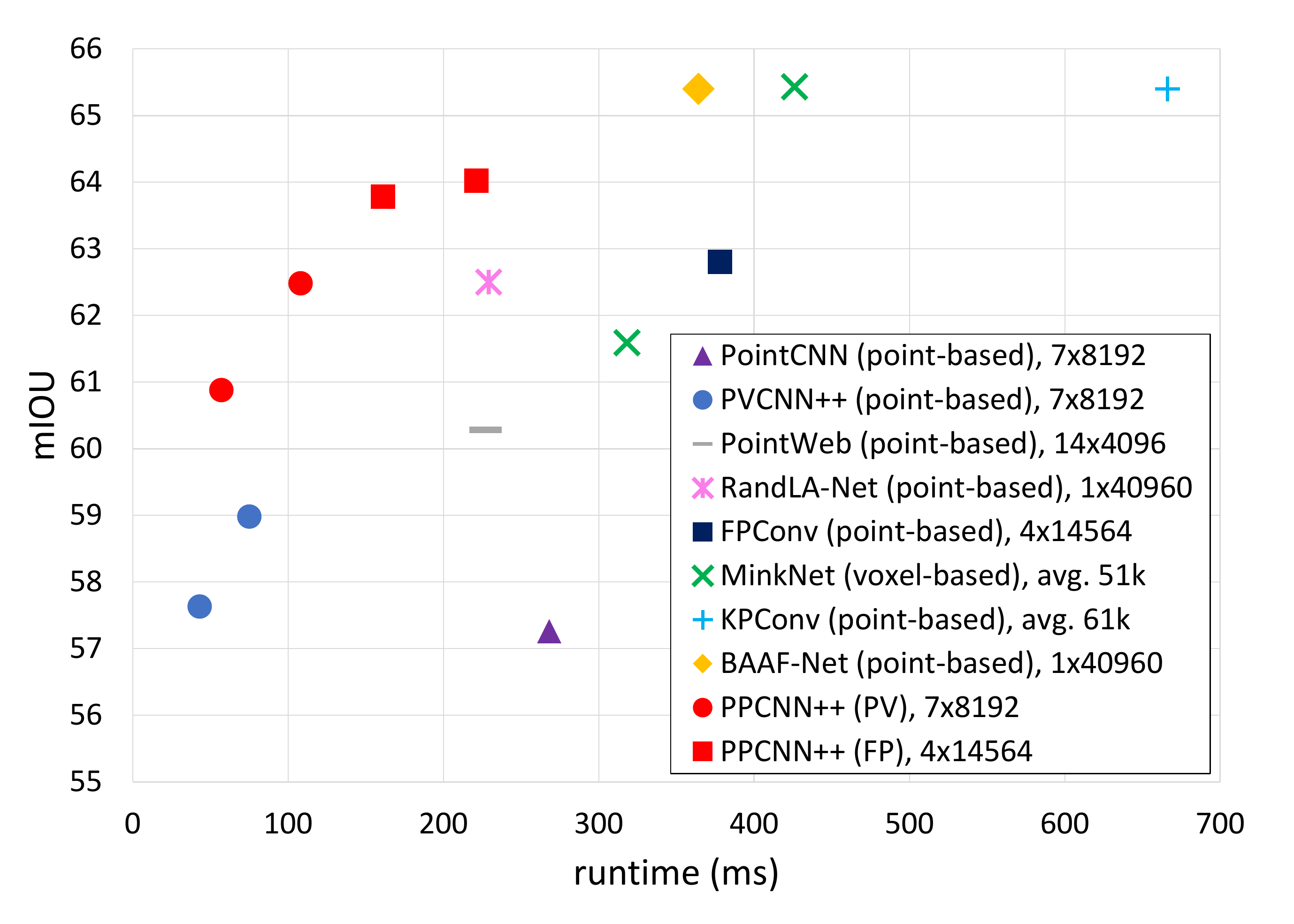}
    \caption{Efficiency comparison on S3DIS indoor scene segmentation (Area-5). ``(PV)'' and ``(FP)'' for PPCNN++ models indicate data pre-processing methods, as explained in Section~\ref{sec:scene_seg_dataset}.}
    \label{fig:graph_s3dis}
\end{figure}

\begin{table}[h]
\centering
\caption{S3DIS (Area-5) results presented in Fig.~\ref{fig:graph_s3dis}}
\begin{adjustbox}{max width=\columnwidth}
\begin{tabular}{|c|c|c|}
\hline
Method & mIoU & Runtime \\ \hhline{|=|=|=|}
PointCNN & 57.26 & 268ms \\
PVCNN++ (0.5C) & 57.63 & 43ms \\
PVCNN++ & 58.98 & 75ms \\
PointWeb & 60.28 & 227ms \\
RandLA-Net & 62.5 & 229ms \\
FPConv & 62.8 & 378ms \\
Mink16UNet14 & 61.59 & 318ms \\
Mink16UNet34 & 65.43 & 426ms\\
BAAF-Net & 65.4 & 364ms\\
KPConv (rigid) & 65.4 & 666ms\\\hline
PPCNN++ (z) (PV) & 60.88 & 57ms\\
PPCNN++ (x,y,z,CAF) (PV) & 62.48 & 108ms \\
PPCNN++ (z) (FP) & 63.78 & 161ms\\
PPCNN++ (x,y,z,IWF) (FP) & 64.02 & 221ms\\ \hline
\end{tabular}
\end{adjustbox}
\label{tab:s3dis}
\end{table}

\begin{table}[h]
\centering
\caption{S3DIS 6-fold cross validation results}
\begin{adjustbox}{max width=\textwidth}
\begin{tabular}{|c|c|c|c|}
\hline
Method & mIoU \\ \hhline{|=|=|}
PointCNN & 65.4 \\
PointWeb & 66.7 \\
FPConv & 68.7 \\
RandLA-Net & 70.0 \\
KPConv (rigid) & 69.6 \\\
BAAF-Net & 72.2 \\ \hline
PPCNN++ (x,y,z,IWF) (FP) & 67.5 \\ \hline
\end{tabular}
\end{adjustbox}
\label{tab:s3dis_6fold}
\end{table}

\begin{figure}[h]
\centering
\includegraphics[width=\columnwidth]{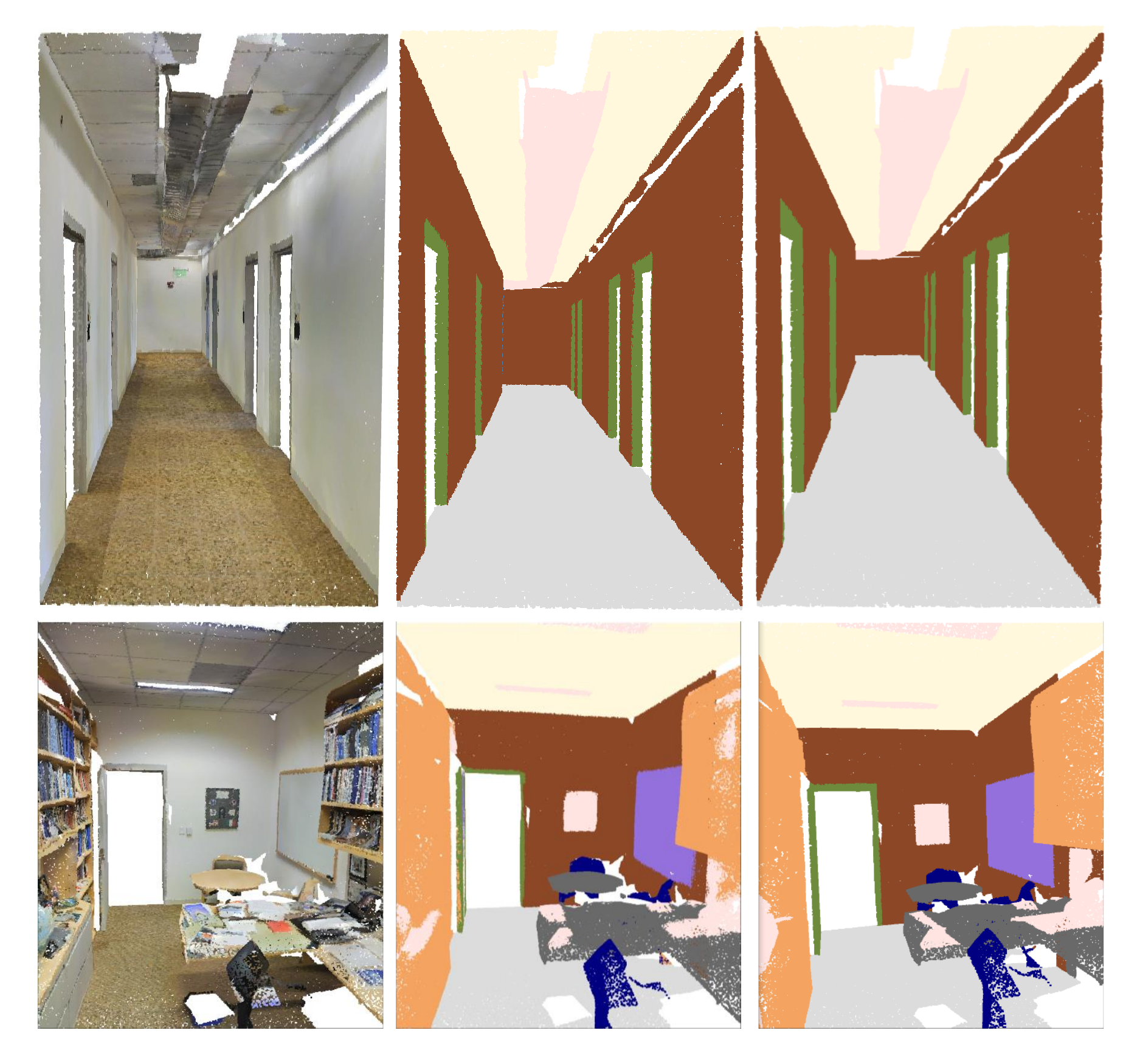}
\caption{Visualized predictions by PPCNN++ (middle) on S3DIS, presented along with the corresponding input point cloud (left) and the ground truth (right).}
\label{fig:vis_s3dis}
\end{figure}

\subsubsection{Results}
\label{sec:scene_seg_results}

The indoor scene segmentation results are shown in Table~\ref{tab:s3dis} and visually presented in Fig.~\ref{fig:graph_s3dis}. In the graph, we provide the number of input points for each method in the legend. The graph shows that PPCNN++ shows better efficiency than any other compared methods. Some state-of-the-art methods, such as BAAF-Net~\cite{qiu2021semantic}, MinkowskiNet~\cite{choy20194d}, and KPConv~\cite{thomas2019kpconv}, show better performance than PPCNN++ with significantly slower inference speed. For example, BAAF-Net, which is the fastest among those models, takes 364ms to process 40k points, while PPCNN++ model with the best mIoU takes 221ms to process 58k points. When compared to RandLA-Net~\cite{hu2020randla}, a point-based model known to be extremely fast, PPCNN++ with FPConv protocol (red square) shows better efficiency trade-off. Overall, the experimental results clearly show that PPCNN++ can perform point cloud segmentation with a small gap of accuracy from the state-of-the-art models, but with a remarkable advantage in inference speed. Additionally, 6-fold cross validation results are shown in Table~\ref{tab:s3dis_6fold}, class-wise mIoUs are shown in Table~\ref{tab:s3dis_class}, and visualized examples of the prediction of PPCNN++ are shown in Fig.~\ref{fig:vis_s3dis}. We present the results reported in each paper, thus only those papers that published 6-fold cross validation results (Table~\ref{tab:s3dis_6fold}) or class-wise mIoUs (Table~\ref{tab:s3dis_class}) are shown.

\begin{table*}[t]
\centering
\caption{Mean IoU for each class for S3DIS (Area-5) (\textbf{Bold}: the best result for each class, \underline{Underlined}: the second best result)}
\begin{adjustbox}{max width=\textwidth}
\begin{tabular}{|c||c||c|c|c|c|c|c|c|c|c|c|c|c|c|}
\hline
Method              & mIoU  & ceiling & floor & wall & beam & column & window & door & table & chair & sofa & bookcase & board & clutter \\ \hhline{|=|=|=|=|=|=|=|=|=|=|=|=|=|=|=|}
PointCNN            & 57.26 & 92.31 & 98.24 & 79.41 & 0.0 & 17.60 & 22.77 & 62.09 & 74.39 & 80.59 & 31.67 & 66.67 & 62.05 & 56.74 \\
SuperPointGraph     & 58.04 & 89.35 & 96.87 & 78.12 & 0.0 & \textbf{42.81} & 48.93 & 61.58 & 84.66 & 75.41 & 69.84 & 52.60 & 2.10 & 52.22 \\
PointWeb            & 60.28 & 91.95 & 98.48 & 79.39 & 0.0 & 21.11 & 59.72 & 34.81 & 76.33 & \textbf{88.27} & 46.89 & 69.30 & 64.91 & 52.46 \\
FPConv              & 62.8  & \textbf{94.6} & 98.5 & 80.9 & 0.0 & 19.1 & \underline{60.1} & 48.9 & 80.6 & \underline{88.0} & 53.2 & 68.4 & \underline{68.2} & 54.9\\
MinkowskiNet32      & 65.35 & 91.75 & \textbf{98.71} & \textbf{86.19} & 0.0 & \underline{34.06} & 48.90 & 62.44 & 89.82 & 81.57 & \underline{74.88} & 47.21 & \textbf{74.44} & \underline{58.57} \\
KPConv (rigid)      & \underline{65.4}  & 92.6 & 97.3 & 81.4 & 0.0 & 16.5 & 54.5 & \textbf{69.5} & \underline{90.1} & 80.2 & 74.6 & 66.4 & 63.7 & 58.1 \\
KPConv (deform)     & \textbf{67.1}  & 92.8 & 97.3 & 82.4 & 0.0 & 23.9 & 58.0 & \underline{69.0} & \textbf{91.0} & 81.5 & \textbf{75.3} & \textbf{75.4} & 66.7 & \textbf{58.9} \\ \hline
PPCNN++ (xyz, IWF)   & 64.02 & \underline{94.04} & \underline{98.49} & \underline{83.73} & 0.0 & 18.58 & \textbf{66.13} & 61.74 & 79.38 & \underline{88.00} & 49.48 & \underline{70.11} & 66.41 & 56.14\\ \hline
\end{tabular}
\end{adjustbox}
\label{tab:s3dis_class}
\end{table*}

\subsection{Object Part Segmentation}
\label{sec:part_seg}

\subsubsection{Dataset}
\label{sec:part_seg_dataset}

In this subsection, we report the object part segmentation performance of PPCNN++ on ShapeNetPart dataset~\cite{yi2016scalable}. Following previous works~\cite{li2018pointcnn,liu2019point}, we evaluated the models using mIoU averaged over 2,874 test models (instance mIoU), after training each model using 14,006 samples from 16 shape categories. Each training sample consists of 2,048 points with the surface normal as extra feature. The experimental settings that are not specified here were borrowed from PVCNN~\cite{liu2019point}.

\begin{table}[h]
\centering
\caption{Network layer configurations of PPCNN++ for ShapeNetPart experiments}
\begin{adjustbox}{max width=\columnwidth}
\begin{tabular}{|c|c|c|}
\hline
\multirow{4}{*}{SA blocks} & [32, 1, 64] & 512, 0.1, 32, (32, 64) \\
& [64, 1, 32] & 128, 0.2, 32, (64, 128) \\
& [128, 1, 16] & 32, 0.4, 32, (128, 256) \\
& None & 16, 0.8, 16, (256, 512) \\ \hline
\multirow{4}{*}{FP blocks} & (256, 256) & [256, 1, 8] \\
& (256, 256) & [256, 1, 16] \\
& (256, 128) & [128, 1, 32] \\
& (128, 64) & [64, 1, 64] \\ \hline
\end{tabular}
\end{adjustbox}
\label{tab:arch_net_shapenet}
\end{table}

\subsubsection{Architecture and Hyperparameters}
\label{sec:part_seg_arch_param}

We train PPCNN++ model with the grid resolution of 64 $\times$ 64 for the first layer of the network. Each sample of ShapeNetPart dataset is a single object, which means that the geometry of point cloud is simpler than that of S3DIS samples. Thus, we used a smaller number of layers in PPCNN++, which is presented in detail in Table~\ref{tab:arch_net_shapenet}.

\begin{table}[h]
\centering
\caption{ShapeNetPart results}
\begin{adjustbox}{max width=\columnwidth}
\begin{tabular}{|c|c|}
\hline
Method          & ins. mIoU \\ \hhline{|=|=|}
PointNet        & 83.7 \\
SyncSpecCNN     & 84.7 \\
PointNet++      & 85.1 \\
3D-GCN          & 85.1 \\
PCNN            & 85.1 \\
DGCNN           & 85.2 \\
SpiderCNN       & 85.3 \\
SPLATNet        & 85.4 \\
SpecGCN         & 85.4 \\
PointConv       & 85.7 \\
PointCNN        & 86.1 \\
PAConv          & 86.1 \\
PVCNN           & 86.2 \\
RS-CNN          & 86.2 \\
InterpCNN       & 86.3 \\
KPConv          & 86.4 \\
DensePoint      & 86.4 \\ \hline
PPCNN++ (x,y,z,IWF) & 86.2 \\ \hline
\end{tabular}
\end{adjustbox}
\label{tab:partseg}
\end{table}

\begin{table*}[h]
\centering
\caption{Mean IoU for each class for ShapeNetPart}
\begin{adjustbox}{max width=\textwidth}
\begin{tabular}{|c||c||c|c|c|c|c|c|c|c|c|c|c|c|c|c|c|c|}
\hline
Method          & mIoU & aero & bag & cap & car & chair & earpho & guitar & knife & lamp & laptop & motor & mug & pistol & rocket & skateb & table \\ \hhline{|=|=|=|=|=|=|=|=|=|=|=|=|=|=|=|=|=|=|}
PointNet        & 83.7 & 83.4 & 78.7 & 82.5 & 74.9 & 89.6 & 73.0 & 91.5 & 85.9 & 80.8 & 95.3 & 65.2 & 93.0 & 81.2 & 57.9 & 72.8 & 80.6 \\
SyncSpecCNN     & 84.7 & 81.6 & 81.7 & 81.9 & 75.2 & 90.2 & 74.9 & \textbf{93.0} & 86.1 & 84.7 & 95.6 & 66.7 & 92.7 & 81.6 & 60.6 & \textbf{82.9} & 82.1 \\
PointNet++      & 85.1 & 82.4 & 79.0 & 87.7 & 77.3 & 90.8 & 71.8 & 91.0 & 85.9 & 83.7 & 95.3 & 71.6 & 94.1 & 81.3 & 58.7 & 76.4 & 82.6 \\
PCNN            & 85.1 & 82.4 & 80.1 & 85.5 & 79.5 & 90.8 & 73.2 & 91.3 & 86.0 & 85.0 & 95.7 & 73.2 & 94.8 & 83.3 & 51.0 & 75.0 & 81.8 \\
DGCNN           & 85.2 & 84.2 & 83.7 & 84.4 & 77.1 & 90.9 & 78.5 & 91.5 & 87.3 & 82.9 & 96.0 & 67.8 & 93.3 & 82.6 & 59.7 & 75.5 & 82.0 \\
SpiderCNN       & 85.3 & 83.5 & 81.0 & 87.2 & 77.5 & 90.7 & 76.8 & 91.1 & 87.3 & 83.3 & 95.8 & 70.2 & 93.5 & 82.7 & 59.7 & 75.8 & 82.8 \\
SPLATNet        & 85.4 & 83.2 & 84.3 & 89.1 & 80.3 & 90.7 & 75.5 & 92.1 & 87.1 & 83.9 & \textbf{96.3} & 75.6 & \textbf{95.8} & 83.8 & 64.0 & 75.5 & 81.8 \\
PointCNN        & 86.1 & 84.1 & \textbf{86.5} & 86.0 & 80.8 & 90.6 & 79.7 & 92.3 & 88.4 & 85.3 & 96.1 & 77.2 & 95.3 & 84.2 & 64.2 & 80.0 & 83.0 \\
PAConv          & 86.1 & 84.3 & 85.0 & 90.4 & 79.7 & 90.6 & 80.8 & 92.0 & \textbf{88.7} & 82.2 & 95.9 & 73.9 & 94.7 & 84.7 & 65.9 & 81.4 & \textbf{84.0} \\
RS-CNN          & 86.2 & 83.5 & 84.8 & 88.8 & 79.6 & 91.2 & 81.1 & 91.6 & 88.4 & \textbf{86.0} & 96.0 & 73.7 & 94.1 & 83.4 & 60.5 & 77.7 & 83.6 \\
KPConv (rigid)  & 86.2 & 83.8 & 86.1 & 88.2 & \textbf{81.6} & 91.0 & 80.1 & 92.1 & 87.8 & 82.2 & 96.2 & 77.9 & 95.7 & \textbf{86.8} & 65.3 & 81.7 & 83.6 \\
KPConv (deform) & \textbf{86.4} & \textbf{84.6} & 86.3 & 87.2 & 81.1 & 91.1 & 77.8 & 92.6 & 88.4 & 82.7 & 96.2 & \textbf{78.1} & \textbf{95.8} & 85.4 & \textbf{69.0} & 82.0 & 83.6 \\
DensePoint      & \textbf{86.4} & 84.0 & 85.4 & 90.0 & 79.2 & 91.1 & \textbf{81.6} & 91.5 & 87.5 & 84.7 & 95.9 & 74.3 & 94.6 & 82.9 & 64.6 & 76.8 & 83.7 \\ \hline
PPCNN++ & 86.2 & 83.7 & 83.7 & \textbf{90.7} & 80.8 & \textbf{91.4} & 76.5 & 91.4 & 88.4 & 84.2 & 86.1 & 76.1 & 95.5 & 83.8 & 61.8 & 75.3 & 83.5 \\ \hline
\end{tabular}
\end{adjustbox}
\label{tab:partseg_class}
\end{table*}

\subsubsection{Results}

Table~\ref{tab:partseg} presents comparison of PPCNN++ with other state-of-the-art methods. The efficiency comparison using runtime and performance can be better analyzed on relatively larger dataset, \ie, with larger number of input points. Since we already compared runtime on S3DIS dataset, we only report the segmentation performance in this section. The results show that PPCNN++ exhibits comparable performance to the state-of-the-art methods.

In Table~\ref{tab:partseg_class}, we present mean IoU of each class compared with recent methods. PPCNN++ model outperforms all the other methods in two classes, and shows comparable instance mIoU (``mIoU'' column) with state-of-the-art point-based methods. In this table, we only include methods that reported class-wise mIoU in the published paper.

\section{Ablation Study}
\label{sec:ablation_study}

In this section, we present analysis on components of PPConv by providing experimental evidence of the effect of each module. The analysis was performed using S3DIS dataset using experimental protocol of PVCNN~\cite{liu2019point}. For every experiment in ablation studies, we trained the model 5 times with different random seeds, and report the averaged performance.

\begin{table}[h]
\centering
\caption{Ablation study on each branch of PPConv module}
\begin{adjustbox}{max width=\columnwidth}
\begin{tabular}{|c|c|c|c|}
\hline
Branch              &   mIoU    &   Runtime \\\hline
No projection branch&   57.90   &   28ms \\
No point branch     &   55.82   &   91ms \\
Both branches       &   60.38   &   95ms \\\hline
\end{tabular}
\end{adjustbox}
\label{tab:ablation_study_branch}
\end{table}

First, we demonstrate the effect of each branch of PPConv module: projection branch and point branch. We trained the model without each branch and report test performances in Table~\ref{tab:ablation_study_branch}. For the model without point branch, feature fusion was performed by adding 3 projection branch features and then applying a single-layer MLP. The results show that removing either branch shows significant performance drop, supporting that each module extracts complementary features to each other.

\begin{table}[h]
\centering
\caption{Ablation study on the number of projection branches}
\begin{adjustbox}{max width=\columnwidth}
\begin{tabular}{|c|c|c|c|}
\hline
Projection axes &   mIoU    &   Runtime \\\hline
z               &   59.07   &   57ms \\
x, z            &   59.97   &   86ms \\
x, y, z         &   60.38   &   95ms \\\hline
\end{tabular}
\end{adjustbox}
\label{tab:ablation_study_proj_axis}
\end{table}

Next, we evaluated PPCNN++ models with different number of projection branches. We trained 3 different models, each with 1 (z), 2 (x,z), and 3 (x,y,z) projection branches. Every model included point branch in this experiment. The results are shown in Table~\ref{tab:ablation_study_proj_axis}, which shows increasing performance as projection branch is added. On the other hand, less projection branch results in faster runtime, which gives possible choices over the trade-off between runtime and performance.

\begin{table}[h]
\centering
\caption{Ablation study on projection methods}
\begin{adjustbox}{max width=\columnwidth}
\begin{tabular}{|c|c|c|}
\hline
Projection methods      & mIoU  & Runtime  \\ \hline
Average features        & 59.30 & 83ms \\
Bilinear Interpolation  & 59.78 & 131ms \\
PointNet-based          & 60.38 & 95ms \\ \hline
\end{tabular}
\end{adjustbox}
\label{tab:ablation_proj_method}
\end{table}

We also evaluated different projection methods to demonstrate their effect on the segmentation performance. PointNet-based projection was mainly used throughout the experiments in the paper. We evaluated simpler projection methods, averaging features and bilinear interpolation, which was explained in Section~\ref{sec:projection}. The results in Table~\ref{tab:ablation_proj_method} present that PointNet-based projection is clearly better than the other methods, justifying the use of PointNet-based projection for segmentation.

\begin{table}[h]
\centering
\caption{Ablation study on the grid resolution}
\begin{adjustbox}{max width=\columnwidth}
\begin{tabular}{|c|c|c|}
\hline
Grid resolution & mIoU & Runtime \\ \hline
32 & 59.44 & 93ms \\
48 & 59.82 & 93ms \\
64 & 60.38 & 95ms \\
96 & 59.56 & 98ms \\ \hline
\end{tabular}
\end{adjustbox}
\label{tab:ablation_grid_resolution}
\end{table}

For the grid resolution during projection, we tested a few choices in order to determine the optimal value. We evaluated four different resolutions: 32$\times$32, 48$\times$48, 64$\times$64, and 96$\times$96. Through experimental results, which is shown in Table~\ref{tab:ablation_grid_resolution}, we observed that 64$\times$64 shows the best performance and chose this value as the default experimental setting in this paper.

\begin{table}[h]
\centering
\caption{Ablation study on the 2D convolutional module}
\begin{adjustbox}{max width=\columnwidth}
\begin{tabular}{|c|c|c|}
\hline
2D convolutional module & mIoU & Runtime \\ \hline
[Conv-BN-LReLU] $\times$2 & 59.82 & 88ms \\
+ Residual connection & 60.03 & 88ms \\
+ Squeeze-and-Excitation module & 60.38 & 95ms \\\hline
\end{tabular}
\end{adjustbox}
\label{tab:ablation_2d_conv}
\end{table}

Then, we present experimental results of variations in 2D convolutional modules. As 2D convolutional modules deal with projected feature maps, which are of the same shape as general image feature maps, any techniques that can be combined with 2D convolutions can be used. We test two of the most popular modules: residual connection and Squeeze-and-Excitation~\cite{hu2018squeeze} module. We trained three different models: without any of the two components, with only residual connection, and with both modules. The segmentation performance, as shown in Table~\ref{tab:ablation_2d_conv}, increases as each module is added to the 2D convolutional module.

\begin{table}[h]
\centering
\caption{Ablation study on feature fusion methods}
\begin{adjustbox}{max width=\columnwidth}
\begin{tabular}{|c|c|c|c|}
\hline
Fusion methods              & mIoU  & Runtime \\\hline
Addition + Concat-MLP       & 60.38 & 95ms \\
Importance-Weighted Fusion  & 61.18 & 106ms \\
Context-Aware Fusion        & 61.89 & 108ms \\\hline
\end{tabular}
\end{adjustbox}
\label{tab:ablation_study_fusion}
\end{table}

Finally, we trained models with different fusion strategies to demonstrate the effectiveness of proposed fusion modules. The default fusion adds 3 projection branch features and then concatenate it with the point branch feature. Then, a single-layer MLP is applied to produce the output feature with the pre-defined channel dimension. We proposed 2 different fusion modules in Section~\ref{sec:feature_fusion}, and provide the experimental results in Table~\ref{tab:ablation_study_fusion}. The results show that both fusion modules contribute to performance improvement over the default method, while runtime is also increased due to the additional MLPs and matrix computations. Context-Aware Fusion module shows noticeably larger improvement while increase of runtime is comparable to that of Importance-Weighted Fusion.

\section{Conclusion}
We proposed PPConv, an efficient local aggregation module for 3D point cloud processing. The experimental observations demonstrated that efficient 2D convolutional modules can facilitate faster computation while keeping performance not far behind from that of 3D convolutional modules. PPConv shows remarkable efficiency in terms of the trade-off between inference time and segmentation performance, providing an efficient alternative for point-based convolutional models. PPConv can process both object part segmentation and scene segmentation under appropriate network architectures, thus can be used as a generic building block for 3D point cloud processing networks.

\begin{IEEEbiography}[{\includegraphics[width=1in,height=1.25in,clip,keepaspectratio]{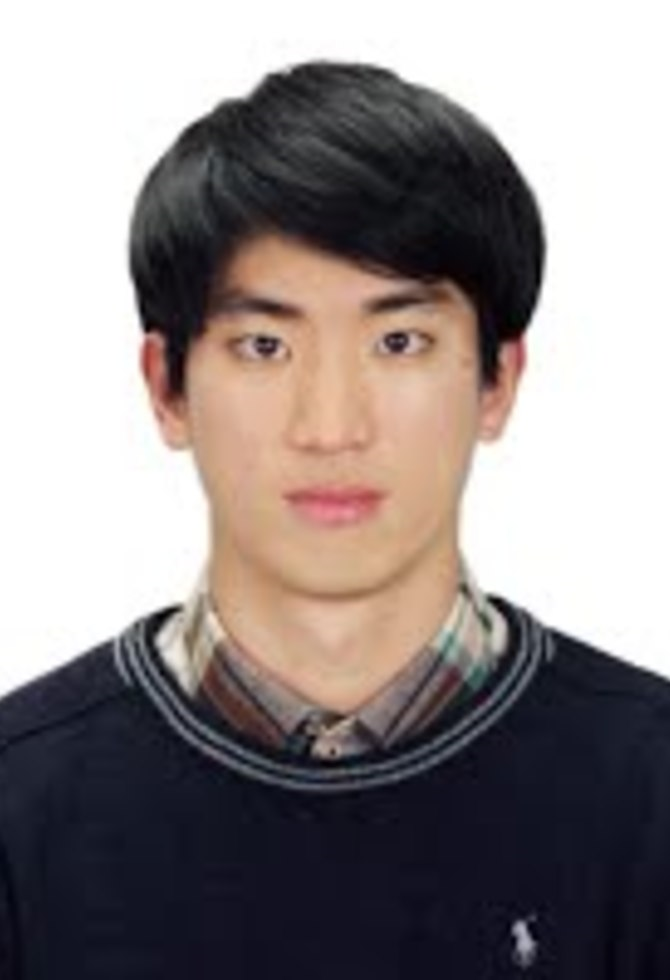}}]{Pyunghwan Ahn} received the B.S. and M.S. degrees in electrical engineering from Korea Advanced Institute of Science and Technology (KAIST), Daejeon, South Korea, in 2014 and 2017, respectively. He is currently pursuing the Ph.D. degree in electrical engineering in KAIST. His research interests include computer vision and 3D deep learning.
\end{IEEEbiography}

\begin{IEEEbiography}[{\includegraphics[width=1in,height=1.25in,clip,keepaspectratio]{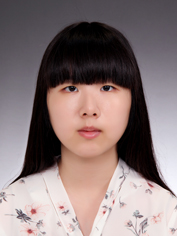}}]{Juyoung Yang} received the B.S. degree in school of integrated technology from Yonsei University, Seoul, South Korea, in 2017. She is currently pursuing the integrated Ph.D. degree in electrical engineering from Korea Advanced Institute of Science and Technology (KAIST), Daejeon, South Korea. Her research interests include computer vision, deep learning, and machine learning.
\end{IEEEbiography}

\begin{IEEEbiography}[{\includegraphics[width=1in,height=1.25in,clip,keepaspectratio]{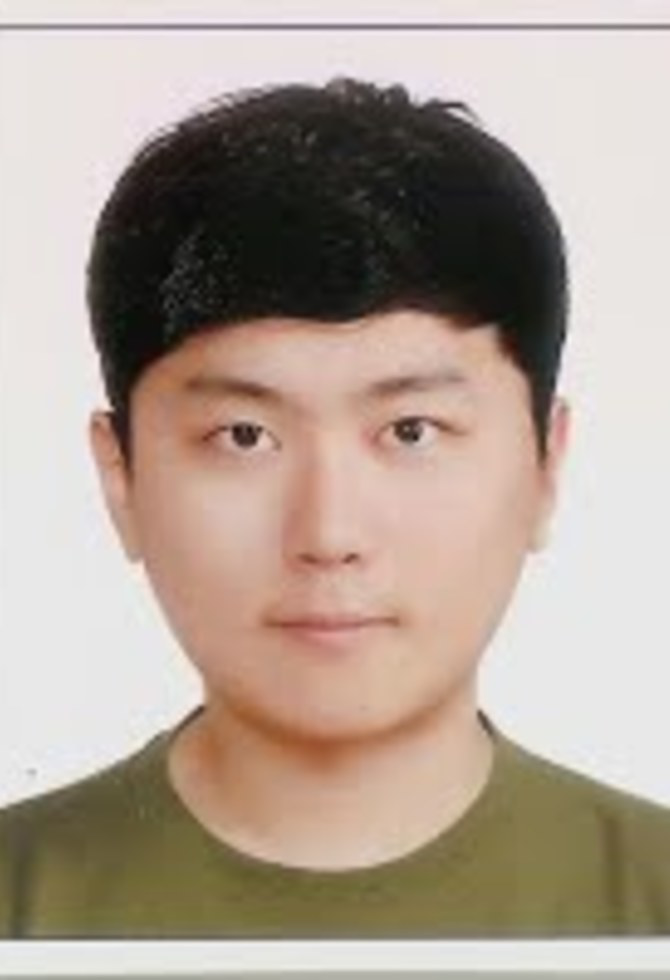}}]{Chanho Lee} received the B.S. and M.S. degrees in electrical engineering from the Korea Advanced Institute of Science and Technology (KAIST), Daejeon, South Korea, in 2016 and 2019, respectively, where he is currently pursuing the Ph.D. degree in electrical engineering. His research interests include computer vision and deep learning.
\end{IEEEbiography}

\begin{IEEEbiography}[{\includegraphics[width=1in,height=1.25in,clip,keepaspectratio]{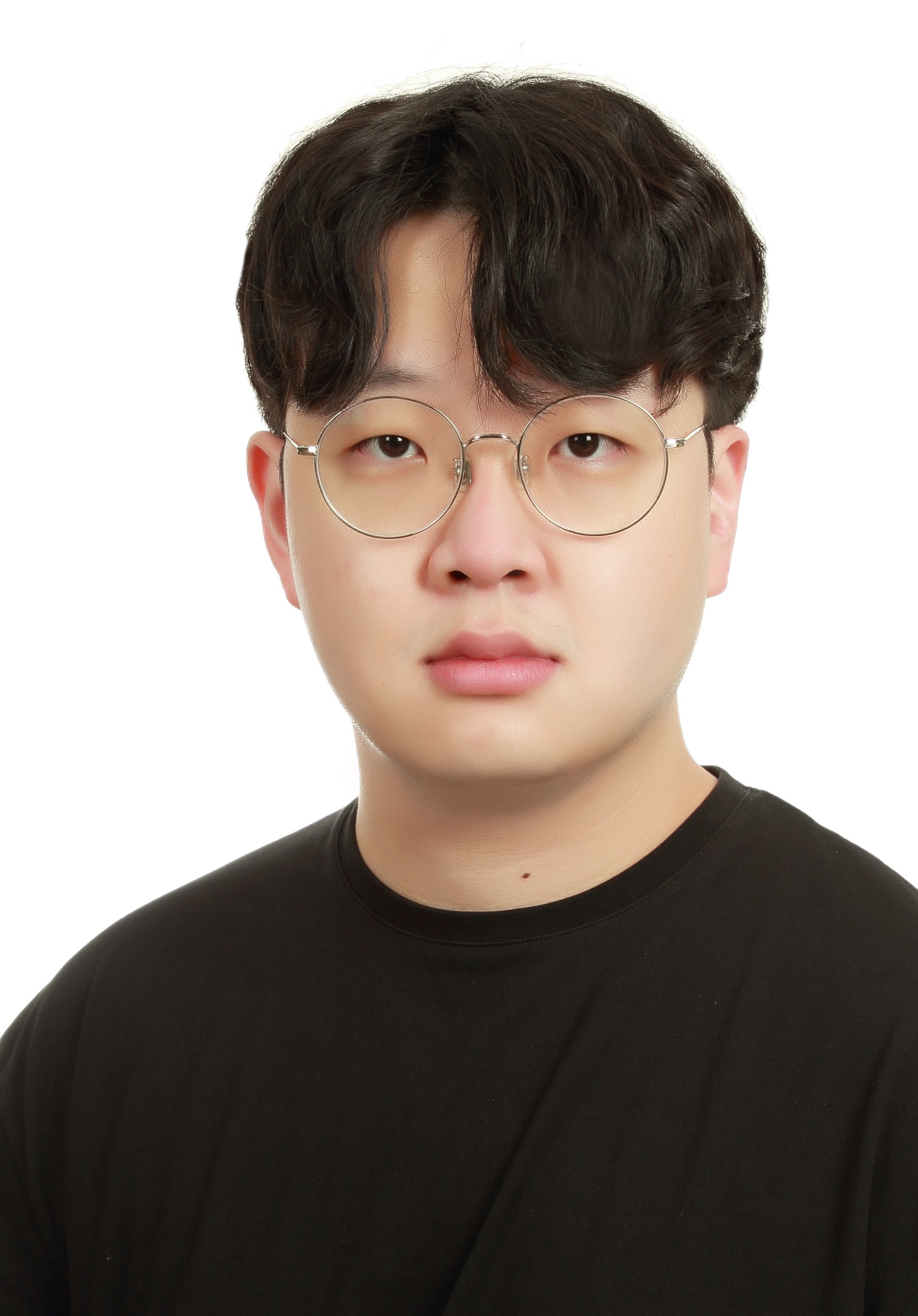}}]{Eojindl Yi} received the B.S. degree in electrical engineering from Hanyang University, Seoul, South Korea, in 2019. He is currently pursuing the integrated Ph.D. degree in electrical engineering from Korea Advanced Institute of Science and Technology (KAIST), Daejeon, South Korea. His research interests include computer vision, deep learning, and machine learning.
\end{IEEEbiography}

\begin{IEEEbiography}[{\includegraphics[width=1in,height=1.25in,clip,keepaspectratio]{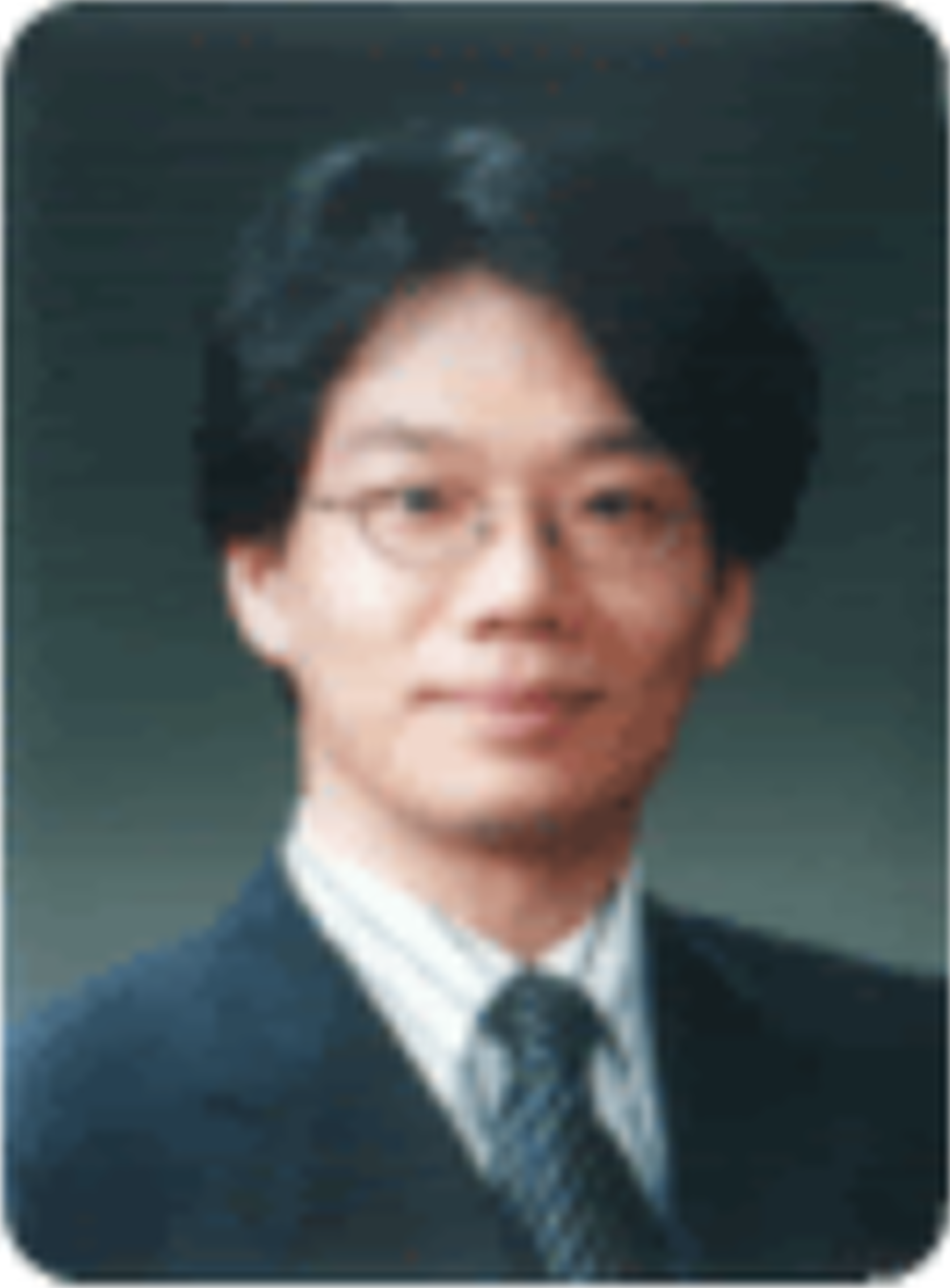}}]{Junmo Kim} (Member, IEEE) received the B.S. degree from Seoul National University, Seoul, South Korea, in 1998, and the M.S. and the Ph.D. degrees from Massachusetts Institute of Technology (MIT), Cambridge, in 2000 and 2005, respectively. From 2005 to 2009, he was a Research Staff Member with Samsung Advanced Institute of Technology (SAIT), South Korea. He joined the faculty of Korea Advanced Institute of Science and Technology (KAIST), in 2009, where he is currently an Associate Professor in School of Electrical Engineering. His research interests include image processing, computer vision, statistical signal processing, machine learning, and information theory.
\end{IEEEbiography}

\EOD


\begin{thebibliography}{00}

\bibitem{armeni20163d} I. Armeni, O. Sener, A. R. Zamir, H. Jiang, I. Brilakis, M. Fischer, and S. Savarese, ``3d semantic parsing of large-scale indoor spaces,'' in \textit{Proc. IEEE Conference on Computer Vision and Pattern Recognition (CVPR)}, Las Vegas, Nevada, USA, 2016, pp. 1534-1543.

\bibitem{atzmon2018point} M. Atzmon, H. Maron, and Y. Lipman, ``Point convolutional neural networks by extension operators,'' 2018, arXiv:1803.10091. [Online]. Available: https://arxiv.org/abs/1803.10091

\bibitem{cheng20212} R. Cheng, R. Razani, E. Taghavi, E. Li, and B. Liu, ``2-S3Net: Attentive feature fusion with adaptive feature selection for sparse semantic segmentation network,'' in \textit{Proc. IEEE/CVF Conference on Computer Vision and Pattern Recognition (CVPR)}, Virtual, 2021, pp. 12547-12556.

\bibitem{choy20194d} C. Choy, J. Gwak, and S. Savarese, ``4d spatio-temporal convnets: Minkowski convolutional neural networks,'' in \textit{Proc. IEEE Conference on Computer Vision and Pattern Recognition (CVPR)}, Long Beach, CA, USA, 2019, pp. 3075-3084.

\bibitem{dosovitskiy2020image} A. Dosovitskiy, L. Beyer, A. Kolesnikov, D. Weissenborn, X. Zhai, T. Unterthiner, M. Dehghani, M. Minderer, G. Heigold, S. Gelly, and J. Uszkoreit, ``An image is worth 16x16 words: Transformers for image recognition at scale,'' 2020, arXiv:2010.11929. [Online]. Available: https://arxiv.org/abs/2010.11929

\bibitem{duan2019structural} Y. Duan, Y. Zheng, J. Lu, J. Zhou, and Q. Tian, ``Structural relational reasoning of point clouds,'' in \textit{Proc. IEEE Conference on Computer Vision and Pattern Recognition (CVPR)}, Long Beach, CA, USA, 2019, pp. 949-958.

\bibitem{fan2021scf} S. Fan, Q. Dong, F. Zhu, Y. Lv, P. Ye, and F. Y. Wang, ``SCF-Net: Learning Spatial Contextual Features for Large-Scale Point Cloud Segmentation,'' in \textit{Proc. IEEE Conference on Computer Vision and Pattern Recognition (CVPR)}, Virtual, 2021, pp. 14504-14513.

\bibitem{gong2021omni} J. Gong, J. Xu, X. Tan, H. Song, Y. Qu, Y. Xie, and L. Ma, ``Omni-supervised Point Cloud Segmentation via Gradual Receptive Field Component Reasoning,'' in \textit{Proc. IEEE Conference on Computer Vision and Pattern Recognition (CVPR)}, Virtual, 2021, pp. 11673-11682.

\bibitem{graham20183d} B. Graham, M. Engelcke, and L. Van Der Maaten, ``3d semantic segmentation with submanifold sparse convolutional networks,'' in \textit{Proc. IEEE Conference on Computer Vision and Pattern Recognition (CVPR)}, Salt Lake City, Utah, USA, 2018, pp. 9224-9232.

\bibitem{he2016deep} K. He, X. Zhang, S. Ren, and J. Sun, ``Deep residual learning for image recognition,'' in \textit{Proc. IEEE Conference on Computer Vision and Pattern Recognition (CVPR)}, Las Vegas, Nevada, USA, 2016, pp. 770-778.

\bibitem{hu2018squeeze} J. Hu, L. Shen, and G. Sun, ``Squeeze-and-excitation networks,'' in \textit{Proc. IEEE Conference on Computer Vision and Pattern Recognition (CVPR)}, Salt Lake City, Utah, USA, 2018, pp. 7132-7141.

\bibitem{hu2020randla} Q. Hu, B. Yang, L. Xie, S. Rosa, Y. Guo, Z. Wang, N. Trigoni, and A. Markham, ``Randla-net: Efficient semantic segmentation of large-scale point clouds,'' in \textit{Proc. IEEE Conference on Computer Vision and Pattern Recognition (CVPR)}, Virtual, 2020, pp. 11108-11117.

\bibitem{ioffe2015batch} S. Ioffe and C. Szegedy, ``Batch normalization: Accelerating deep network training by reducing internal covariate shift,'' in \textit{Proc. International conference on machine learning (ICML)}, Lille, France, 2015, pp. 448-456.

\bibitem{landrieu2018large} L. Landrieu and M. Simonovsky, ``Large-scale point cloud semantic segmentation with superpoint graphs,'' in \textit{Proc. IEEE Conference on Computer Vision and Pattern Recognition (CVPR)}, Salt Lake City, Utah, USA, 2018, pp. 4558-4567.

\bibitem{lang2019pointpillars} A. H. Lang, S. Vora, H. Caesar, L. Zhou, J. Yang, and O. Beijbom, ``Pointpillars: Fast encoders for object detection from point clouds,''  in \textit{Proc. IEEE Conference on Computer Vision and Pattern Recognition (CVPR)}, Long Beach, CA, USA, 2019, pp. 12697-12705.

\bibitem{lawin2017deep} F. J. Lawin, M. Danelljan, P. Tosteberg, G. Bhat, F. S. Khan, and M. Felsberg, ``Deep projective 3D semantic segmentation,'' in \textit{Proc. International Conference on Computer Analysis of Images and Patterns (CAIP)}, Ystad, Sweden, 2017, pp. 95-107.

\bibitem{li2020end} L. Li, S. Zhu, H. Fu, P. Tan, and C. L. Tai, ``End-to-end learning local multi-view descriptors for 3d point clouds,'' in \textit{Proc. IEEE Conference on Computer Vision and Pattern Recognition (CVPR)}, Virtual, 2020, pp. 1919-1928.

\bibitem{li2018pointcnn} Y. Li, R. Bu, M. Sun, W. Wu, X. Di, and B. Chen, ``Pointcnn: Convolution on x-transformed points,'' in \textit{Proc. Conference on Neural Information Processing Systems}, Montreal, Canada, 2018, 820-830.

\bibitem{lin2020fpconv} Y. Lin, Z. Yan, H. Huang, D. Du, L. Liu, S. Cui, and X. Han, ``Fpconv: Learning local flattening for point convolution,'' in \textit{Proc. IEEE Conference on Computer Vision and Pattern Recognition (CVPR)}, Virtual, 2020, pp. 4293-4302.

\bibitem{lin2021learning} Z. H. Lin, S. Y. Huang, and Y. C. F. Wang, ``Learning of 3D Graph Convolution Networks for Point Cloud Analysis,'' \textit{IEEE Transactions on Pattern Analysis and Machine Intelligence}, Early Access, Feb. 2021.

\bibitem{lionar2021dynamic} S. Lionar, D. Emtsev, D. Svilarkovic, and S. Peng, ``Dynamic Plane Convolutional Occupancy Networks,'' in \textit{Proc. IEEE/CVF Winter Conference on Applications of Computer Vision (WACV)}, Virtual, 2021, pp. 1829-1838.

\bibitem{liu2019densepoint} Y. Liu, B. Fan, G. Meng, J. Lu, S. Xiang, and C. Pan, ``Densepoint: Learning densely contextual representation for efficient point cloud processing,'' in \textit{Proc. IEEE Conference on Computer Vision and Pattern Recognition (CVPR)}, Long Beach, CA, USA, 2019, pp. 5239-5248.

\bibitem{liu2019relation} Y. Liu, B. Fan, S. Xiang, and C. Pan, ``Relation-shape convolutional neural network for point cloud analysis,'' in \textit{Proc. IEEE Conference on Computer Vision and Pattern Recognition (CVPR)}, Long Beach, CA, USA, 2019, pp. 8895-8904.

\bibitem{liu2019point} Z. Liu, H. Tang, Y. Lin, and S. Han, ``Point-Voxel CNN for Efficient 3D Deep Learning,'' in \textit{Proc. Conference on Neural Information Processing Systems}, Vancouver, Canada, 2019, 965-975.

\bibitem{lu2021cga} T. Lu, L. Wang, and G. Wu, ``CGA-Net: Category Guided Aggregation for Point Cloud Semantic Segmentation,'' in \textit{Proc. IEEE Conference on Computer Vision and Pattern Recognition (CVPR)}, Virtual, 2021, pp. 11693-11702.

\bibitem{mao2019interpolated} J. Mao, X. Wang, and H. Li, ``Interpolated convolutional networks for 3d point cloud understanding,'' in \textit{Proc. IEEE Conference on Computer Vision and Pattern Recognition (CVPR)}, Long Beach, CA, USA, 2019, pp. 1578-1587.

\bibitem{maturana2015voxnet} D. Maturana and S. Scherer, ``Voxnet: A 3d convolutional neural network for real-time object recognition,'' in \textit{Proc. IEEE/RSJ International Conference on Intelligent Robots and Systems (IROS)}, Hamburg, Germany, 2015, pp. 922-928.

\bibitem{peng2020convolutional} S. Peng, M. Niemeyer, L. Mescheder, M. Pollefeys, and A. Geiger, ``Convolutional occupancy networks,'' in \textit{Proc. European Conference on Computer Vision (ECCV)}, Virtual, 2020, part 3, 16, pp. 523-540.

\bibitem{qi2017pointnet} C. R. Qi, H. Su, K. Mo, and L. J. Guibas, ``Pointnet: Deep learning on point sets for 3d classification and segmentation,'' in \textit{Proc. IEEE Conference on Computer Vision and Pattern Recognition (CVPR)}, Honolulu, HI, USA, 2017, pp. 652-660.

\bibitem{qi2017pointnet++} C. R. Qi, L. Yi, H. Su, and L. J. Guibas, ``PointNet++: Deep Hierarchical Feature Learning on Point Sets in a Metric Space,'' in \textit{Proc. Conference on Neural Information Processing Systems}, Long Beach, CA, USA, 2017.

\bibitem{qiu2021semantic} S. Qiu, S. Anwar, and N. Barnes, ``Semantic Segmentation for Real Point Cloud Scenes via Bilateral Augmentation and Adaptive Fusion,'' in \textit{Proc. IEEE Conference on Computer Vision and Pattern Recognition (CVPR)}, Virtual, 2021, pp. 1757-1767.

\bibitem{roveri2018network} R. Roveri, L. Rahmann, C. Oztireli, and M. Gross, ``A network architecture for point cloud classification via automatic depth images generation,'' in \textit{Proc. IEEE Conference on Computer Vision and Pattern Recognition (CVPR)}, Salt Lake City, Utah, USA, 2018, pp. 4176-4184.

\bibitem{su2018splatnet} H. Su, V. Jampani, D. Sun, S. Maji, E. Kalogerakis, M. H. Yang, and J. Kautz, ``Splatnet: Sparse lattice networks for point cloud processing,'' in \textit{Proc. IEEE Conference on Computer Vision and Pattern Recognition (CVPR)}, Salt Lake City, Utah, USA, 2018, pp. 2530-2539.

\bibitem{su2015multi} H. Su, S. Maji, E. Kalogerakis, and E. Learned-Miller, ``Multi-view convolutional neural networks for 3d shape recognition,'' in \textit{Proc. IEEE international Conference on Computer Vision (ICCV)}, Santiago, Chile, 2015, pp. 945-953.

\bibitem{tang2020searching} H. Tang, Z. Liu, S. Zhao, Y. Lin, J. Lin, H. Wang, and S. Han, ``Searching efficient 3d architectures with sparse point-voxel convolution,'' in \textit{Proc. European Conference on Computer Vision (ECCV)}, Virtual, 2020, pp. 685-702.

\bibitem{tatarchenko2018tangent} M. Tatarchenko, J. Park, V. Koltun, and Q. Y. Zhou, ``Tangent convolutions for dense prediction in 3d,'' in \textit{Proc. IEEE Conference on Computer Vision and Pattern Recognition (CVPR)}, Salt Lake City, Utah, USA, 2018, pp. 3887-3896.

\bibitem{thomas2019kpconv} H. Thomas, C. R. Qi, J. E. Deschaud, B. Marcotegui, F. Goulette, and L. J. Guibas, ``Kpconv: Flexible and deformable convolution for point clouds,'' in \textit{Proc. IEEE Conference on Computer Vision and Pattern Recognition (CVPR)}, Long Beach, CA, USA, 2019, pp. 6411-6420.

\bibitem{touvron2021training} H. Touvron, M. Cord, M. Douze, F. Massa, A. Sablayrolles, and H. Jégou, ``Training data-efficient image transformers \& distillation through attention,'' in \textit{Proc. International conference on machine learning (ICML)}, Virtual, pp. 10347-10357.

\bibitem{wang2018local} C. Wang, B. Samari, and K. Siddiqi, ``Local spectral graph convolution for point set feature learning,'' in \textit{Proc. European Conference on Computer Vision (ECCV)}, Munich, Germany, 2018, pp. 52-66.

\bibitem{wang2017cnn} P. S. Wang, Y. Liu, Y. X. Guo, C. Y. Sun, and X. Tong, ``O-cnn: Octree-based convolutional neural networks for 3d shape analysis,'' \textit{ACM Transactions On Graphics (TOG)}, vol. 36, no. 4, pp. 1-11, Jul. 2017.

\bibitem{wang2020pillar} Y. Wang, A. Fathi, A. Kundu, D. A. Ross, C. Pantofaru, T. Funkhouser, and J. Solomon, ``Pillar-based object detection for autonomous driving,'' in \textit{Proc. European Conference on Computer Vision (ECCV)}, Virtual, 2020, part 22, 16, pp. 18-34.

\bibitem{wang2019dynamic} Y. Wang, Y. Sun, Z. Liu, S. E. Sarma, M. M. Bronstein, and J. M. Solomon, ``Dynamic graph cnn for learning on point clouds,'' \textit{ACM Transactions On Graphics (TOG)}, vol. 38, no. 5, pp. 1-12, Oct. 2019.

\bibitem{wu2019pointconv} W. Wu, Z. Qi, and L. Fuxin, ``Pointconv: Deep convolutional networks on 3d point clouds,'' in \textit{Proc. IEEE Conference on Computer Vision and Pattern Recognition (CVPR)}, Long Beach, CA, USA, 2019, pp. 9621-9630.

\bibitem{xu2021paconv} M. Xu, R. Ding, H. Zhao, and X. Qi, ``PAConv: Position Adaptive Convolution with Dynamic Kernel Assembling on Point Clouds,'' in \textit{Proc. IEEE Conference on Computer Vision and Pattern Recognition (CVPR)}, Virtual, 2021, pp. 3173-3182.

\bibitem{xu2020grid} Q. Xu, X. Sun, C. Y. Wu, P. Wang, and U. Neumann, ``Grid-gcn for fast and scalable point cloud learning,'' in \textit{Proc. IEEE Conference on Computer Vision and Pattern Recognition (CVPR)}, Virtual, 2020, pp. 5661-5670.

\bibitem{xu2018spidercnn} Y. Xu, T. Fan, M. Xu, L. Zeng, and Y. Qiao, ``Spidercnn: Deep learning on point sets with parameterized convolutional filters,'' in \textit{Proc. European Conference on Computer Vision (ECCV)}, Munich, Germany, 2018, pp. 87-102.

\bibitem{yang2020pbp} J. Yang, C. Lee, P. Ahn, H. Lee, E. Yi, and J. Kim, ``PBP-Net: Point Projection and Back-Projection Network for 3D Point Cloud Segmentation,'' in \textit{Proc. IEEE/RSJ International Conference on Intelligent Robots and Systems (IROS)}, Virtual, 2020, pp. 8469-8475.

\bibitem{yi2016scalable} L. Yi, V. G. Kim, D. Ceylan, I. C. Shen, M. Yan, H. Su, A. Sheffer, and L. Guibas, ``A scalable active framework for region annotation in 3d shape collections,'' \textit{ACM Transactions On Graphics (TOG)}, vol. 35, no. 6, pp. 1-12, Nov. 2016.

\bibitem{yi2017syncspeccnn} L. Yi, H. Su, X. Guo, and L. J. Guibas, ``Syncspeccnn: Synchronized spectral cnn for 3d shape segmentation,'' in \textit{Proc. IEEE Conference on Computer Vision and Pattern Recognition (CVPR)}, Honolulu, HI, USA, 2017, pp. 2282-2290.

\bibitem{zhang2020deep} F. Zhang, J. Fang, B. Wah, and P. Torr, ``Deep fusionnet for point cloud semantic segmentation,'' in \textit{Proc. European Conference on Computer Vision (ECCV)}, Virtual, 2020, part 26, 16, pp. 644-663.

\bibitem{zhao2019pointweb} H. Zhao, L. Jiang, C. W. Fu, and J. Jia, ``Pointweb: Enhancing local neighborhood features for point cloud processing,'' in \textit{Proc. IEEE Conference on Computer Vision and Pattern Recognition (CVPR)}, Long Beach, CA, USA, 2019, pp. 5565-5573.

\bibitem{zhao2021point} H. Zhao, L. Jiang, J. Jia, P. H. Torr, and V. Koltun, ``Point transformer,'' in \textit{Proc. IEEE international Conference on Computer Vision (ICCV)}, Virtual, 2021, pp. 16259-16268.

\bibitem{zhou2020end} Y. Zhou, P. Sun, Y. Zhang, D. Anguelov, J. Gao, T. Ouyang, J. Guo, J. Ngiam, and V. Vasudevan, ``End-to-end multi-view fusion for 3d object detection in lidar point clouds,'' in \textit{Proc. Conference on Robot Learning}, 2020, pp. 923-932.

\bibitem{zhou2018voxelnet} Y. Zhou and O. Tuzel, ``Voxelnet: End-to-end learning for point cloud based 3d object detection,'' in \textit{Proc. IEEE Conference on Computer Vision and Pattern Recognition (CVPR)}, Salt Lake City, Utah, USA, 2018, pp. 4490-4499.

\end{thebibliography}
\end{document}